\documentclass{isprs} 
\usepackage{hyperref}
\usepackage{subfigure}
\usepackage{setspace}
\usepackage{geometry} 
\usepackage{epstopdf}
\usepackage{todonotes}
\usepackage{rotating}
\usepackage{booktabs}
\usepackage[labelsep=period]{caption}  
\usepackage{breakcites} 
\usepackage{array}
\usepackage{pgfplots}
\usepackage{tikz}
\usepackage{xcolor}
\definecolor{lu1}{RGB}{  249	255	164}
\definecolor{lu2}{RGB}{ 105	255	248}
\definecolor{lu3}{RGB}{ 28	13	255}
\definecolor{lu4}{RGB}{165	165	165}
\definecolor{lu5}{RGB}{ 0	63	0}
\definecolor{lu6}{RGB}{ 0	108	0}
\definecolor{lu7}{RGB}{ 227	255	119}
\definecolor{lu8}{RGB}{ 182	255	5}
\definecolor{lu9}{RGB}{ 147	206	4}
\definecolor{lu10}{RGB}{ 119	167	3}
\definecolor{lu11}{RGB}{ 220	209	89}

\makeatletter
\newenvironment{customlegend}[1][]{%
\begingroup
    \let\addlegendimage=\pgfplots@addlegendimage
    \let\addlegendentry=\pgfplots@addlegendentry

    \pgfplots@init@cleared@structures
    \pgfplotsset{#1}%
}{
    \pgfplots@createlegend
    \endgroup
}
\makeatother

\usepackage{xparse}

\AppendGraphicsExtensions{.tif}

\newcolumntype{v}[1]{>{\raggedright\hspace{0pt}}p{#1}}

\newcommand{\breakingperiod}{%
  \penalty0 
  .\nobreak\hspace{0pt}%
}

\ExplSyntaxOn

\NewDocumentCommand{\longword}{m}
 {
  \texttt
   {
    \seq_set_split:Nnn \l_michael_lw_seq { . } { #1 }
    \seq_use:Nn \l_michael_lw_seq { \breakingperiod }
   }
 }

\ExplSyntaxOff

\usepackage{floatrow}
\DeclareFloatFont{footnotesize}{\footnotesize}
\begin{document}

\title{SEN12MS -- A Curated Dataset of Georeferenced Multi-Spectral Sentinel-1/2 Imagery for Deep Learning and Data Fusion}

\author{
M. Schmitt\textsuperscript{1}, L. H. Hughes\textsuperscript{1}, C. Qiu\textsuperscript{1}, X. X. Zhu\textsuperscript{1,2}}

\address{
	\textsuperscript{1 }Signal Processing in Earth Observation, Technical University of Munich, Munich, Germany\\
	\textsuperscript{2 }Remote Sensing Technology Institute, German Aerospace Center (DLR), Oberpfaffenhofen, Wessling
}


\icwg{}   

\abstract{\textit{This is a pre-print of a paper accepted for publication in the ISPRS Annals of the Photogrammetry, Remote Sensing and Spatial Infor-mation Sciences. Please refer to the original (open access) publication from September 2019.}\\
The availability of curated large-scale training data is a crucial factor for the development of well-generalizing deep learning methods for the extraction of geoinformation from multi-sensor remote sensing imagery. While quite some datasets have already been published by the community, most of them suffer from rather strong limitations, e.g. regarding spatial coverage, diversity or simply number of available samples. Exploiting the freely available data acquired by the Sentinel satellites of the Copernicus program implemented by the European Space Agency, as well as the cloud computing facilities of Google Earth Engine, we provide a dataset consisting of $180{,}662$ triplets of dual-pol synthetic aperture radar (SAR) image patches, multi-spectral Sentinel-2 image patches, and MODIS land cover maps. With all patches being fully georeferenced at a $10$\,m ground sampling distance and covering all inhabited continents during all meteorological seasons, we expect the dataset to support the community in developing sophisticated deep learning-based approaches for common tasks such as scene classification or semantic segmentation for land cover mapping.}


\maketitle


\section{Introduction}\label{sec:Introduction}
\sloppy
The availability of curated annotated datasets is of crucial importance for the development of machine learning models for information retrieval from remote sensing data. While classic \textit{shallow} learning approaches could easily be trained on comparably small datasets such as, e.g., the famous Indian Pines scene \cite{Baumgardner2015}, modern \textit{deep} learning requires large-scale data to reach the desired generalization performance \cite{Zhu2017}. However, computer vision usually deals with conventional photographs of everyday objects, whereas remote sensing data is more versatile and much more difficult to interpret. Therefore, massive databases of labeled imagery such as ImageNet \cite{Deng2009} do not yet exist in the remote sensing domain, although there have been first steps into that direction; a certainly non-exhaustive overview of existing scientific datasets of annotated remote sensing imagery can be found in Tab.~\ref{tab:datasets}. Additional datasets, which were mostly provided in the frame of machine learning competitions and are not described and discussed in scientific papers, can be found in a the private link list of \cite{Rieke2019}.

\begin{table*}[!tbh]
    \centering
    \begin{tabular}{v{2.5cm} v{1.5cm} v{1.8cm} v{2cm} v{3.5cm} l}
        \toprule
        Dataset & Nr. of Images & Image Size & Data Source & Description & Reference\\
        \cmidrule{1-1}  \cmidrule(lr){2-2} \cmidrule(lr){3-3} \cmidrule(lr){4-4} \cmidrule(lr){5-5} \cmidrule{6-6}
        
       \textit{UC Merced Land Use Dataset} & $2{,}100$ & $256\times256$ & Color aerial images & $21$ balanced land use classes & \cite{Yang2010}\\\addlinespace
        
        \textit{SAT-4} & $500{,}000$ & $28\times28$ & Color aerial imagery (R, G, B, NIR) & $4$ agricultural classes over continental USA & \cite{Basu2015}\\\addlinespace
        
        \textit{SAT-6} & $405{,}000$ & $28\times28$ & Color aerial imagery (R, G, B, NIR) & $6$ land cover classes over continental USA & \cite{Basu2015}\\\addlinespace
        
        \textit{Brazilian Coffee Scenes Dataset} & $51{,}004$ & $64\times64$ & SPOT & $2$ imbalanced classes describing non-coffee and coffee & \cite{Penatti2015}\\\addlinespace
        
        \textit{USGS SIRI-WHU} & $1$ & $10{,}000 \times 9{,}000$ & Color aerial image & $4$ classes over Montgomery County, Ohio, USA & \cite{Zhong2015}\\\addlinespace
        
        \textit{Brazilian Cerrado-Savanna Dataset} & $1{,}311$ & $64\times64$ & RapidEye (G, R, NIR) & $4$ imbalanced classes describing Cerrado-Savanna vegetation & \cite{Nogueira2016}\\\addlinespace
        
        \textit{SIRI-WHU} & $200$ & $200\times200$ & Google Earth & $12$ classes describing urban areas in China & \cite{Zhao2016}\\\addlinespace
        
        \textit{Inria Aerial Image Labeling Dataset} & $360$ & $1{,}500\times1{,}500$ & Color aerial imagery & $2$ classes (building / not building) for $10$ cities in Austria and USA & \cite{Maggiori2017}\\\addlinespace
        
        \textit{2017 IEEE GRSS Data Fusion Contest} & $57$ & from $447\times377$ to $1{,}461\times1{,}222$& Sentinel-2, Landsat, OpenStreetMap & $17$ local climate zone classes & \cite{yokoyaOpen}\\ \addlinespace
        
        \textit{DeepGlobe -- Road Extraction} & $8{,}570$ & $1{,}024\times1{,}024$ & Worldview-2/-3, GeoEye-1 (R, G, B) & $1$ target class: roads over India and Thailand& \cite{Demir2018}\\\addlinespace
        
        \textit{DeepGlobe -- Building Detection} & $24{,}586$ & $650\times650$ & Worldview-3 & $1$ target class: buildings in Las Vegas, Shanghai, Paris, Karthoum & \cite{Demir2018}\\\addlinespace
        
        \textit{DeepGlobe\newline -- Land Cover Classification} & $1{,}146$ & $2{,}448\times2{,}448$ & Worldview-2/-3, GeoEye-1 (R, G, B) & $7$ land cover classes & \cite{Demir2018}\\\addlinespace
        
        \textit{DOTA} & $2{,}806$ & $800\times800$\newline to $4{,}000\times4{,}000$ & Color aerial images & $188{,}282$ instances of $15$ object classes, each labeled by a quadriliteral & \cite{Xia2018}\\\addlinespace

        \textit{EuroSAT} & $27{,}000$ & $64\times64$&Sentinel-2 &$10$ classes & \cite{helber2018introducing}\\\addlinespace
        
        \textit{SEN1-2} & $564{,}768$ & $256\times256$&Sentinel-1 and Sentinel-2 &corresponding pairs of SAR (single-pol intensity) and optical (RGB) image pairs without annotations& \cite{Schmitt2018}\\\addlinespace
        
        \textit{38-Cloud} & $17{,}601$ & $384\times384$&Landsat 8 (R, G, B, NIR)&single target class: clouds& \cite{Mohajerani2019}\\\addlinespace
        
        \textit{BigEarthNet} & $590{,}326$ & up to $120\times120$&Sentinel-2 &$43$ Corine Land Cover classes over Europe& \cite{Sumbul2019}\\\addlinespace
        
        \textit{SEN12MS} & $541{,}986$  & $256\times256$ & Sentinel-1, Sentinel-2, MODIS Land Cover & globally distributed; MODIS Land Cover maps can either be used as labels or auxiliary data & \textit{this paper}\\\addlinespace
        \bottomrule
    \end{tabular}
    \caption{Non-exhaustive list of curated datasets for deep learning in remote sensing.}
    \label{tab:datasets}
\end{table*}

In order to support deep learning-related research in the field of remote sensing, we have published the \emph{SEN1-2} dataset in 2018, which is comprised of about $280{,}000$ pairs of corresponding Sentinel-1 SAR and Sentinel-2 optical images \cite{Schmitt2018}. Since \emph{SEN1-2} was mainly intended for bridging the gap between classical computer vision problems and remote sensing, e.g. image-to-image translation tasks, the provided data were strongly simplified:
For Sentinel-1, we just provided vertically polarized (VV) imagery in dB scale, and for Sentinel-2 we reduced the original multi-spectral data tensors to RGB images with adjusted histograms, whereas none of the images came with any form of geolocation information. Based on feedback from the community, with this paper, we now publish a follow-on version of the dataset, which is designed to suit the needs of the remote sensing community. \emph{SEN12MS} contains full multi-spectral information in geocoded imagery and is described in details throughout the remainder of this paper.

\section{The Data Basis}\label{sec:dataBasis}
We exploit freely available satellite(-derived) data to form the basis of the dataset. On the one hand, we make use of SAR and multi-spectral imagery provided by Sentinel-1 and Sentinel-2, respectively. On the other hand, we add land cover information derived from observations acquired by the MODIS system. Details of the three basic data sources are provided in the following.
\subsection{Sentinel-1}
The Sentinel-1 mission \cite{Torres2012} consists of currently two polar-orbiting satellites, equipped with C-band SAR sensors, which enables them to acquire imagery regardless of the weather.
Sentinel-1 works in a pre-programmed operation mode to avoid conflicts and to produce a consistent long-term data archive built for applications based on long time series. Depending on which SAR imaging mode is used, resolutions down to $5$\,m with a wide coverage of up to $400$\,km can be achieved. Furthermore, Sentinel-1 provides dual polarization capabilities and very short revisit times of about 1 week at the equator. Since highly precise spacecraft positions and attitudes are combined with the high accuracy of the range-based SAR imaging principle, Sentinel-1 images come with high out-of-the-box geolocation accuracy \cite{Schubert2015}.   

For the Sentinel-1 images in the \emph{SEN12MS} dataset, again ground-range-detected (GRD) products acquired in the most frequently available interferometric wide swath (IW) mode were used. These images contain the $\sigma^0$ backscatter coefficient in dB scale for every pixel at a pixel spacing of $5$\,m in azimuth and $20$\,m in range. In order to exploit the full potential of Sentinel-1 data, \emph{SEN12MS} contains both VV and VH polarized images. 

For precise ortho-rectification, restituted orbit information was combined with the $30$\,m-SRTM-DEM or the ASTER DEM for high latitude regions where SRTM is not available.
As for the \emph{SEN1-2} dataset, we intend to leave any further pre-processing, e.g. speckle filtering, to the end user and do not manipulate the data any further.

\subsection{Sentinel-2}
The Sentinel-2 mission \cite{Drusch2012} currently comprises two identical polar-orbiting satellites in the same orbit, phased at $180^\circ$ to each other. The mission is meant to provide continuity for multi-spectral imagery of the SPOT and LANDSAT kind, which have provided information about the land surfaces of our Earth for many decades. With its wide swath width of up to 290~km and its high revisit time of 5 days at the equator (based on two satellites) under cloud-free conditions, the Sentinel-2 mission is specifically well-suited to vegetation monitoring within the growing season. 

For the \emph{SEN12MS} dataset, we provide the full multi-spectral image cubes as extracted from the original, precisely georeferenced Sentinel-2 granules. The only manipulation we carried out was to implement a sophisticated mosaicking workflow to avoid the download of cloud-affected images (cf.~Section~\ref{sec:mosaic}). 
 
\subsection{MODIS Land Cover}
MODIS (the Moderate Resolution Imaging Spectroradiometer) is the main instrument on board of the Terra and Aqua satellites. Terra's orbit around the Earth is timed so that it passes from north to south across the equator in the morning, while Aqua passes south to north over the equator in the afternoon. Terra MODIS and Aqua MODIS acquisitions cover the whole Earth with an approximately daily revisit frequency -- at a band-dependent resolution of $250$\,m to $1000$\,m. Based on calibrated MODIS reflectance data, hierarchical classification following the land cover classification system (LCCS) scheme, and sophisticated post-processing for class-specific refinement incorporating prior knowledge, auxiliary information and temporal regularization based on a Markov random field, annually updated global land cover maps for the years 2001--2016 are provided as MCD12Q1 V6 dataset at a ground sampling distance of $500$\,m \cite{SullaMenashe2019}. 
To add land cover information to the Sentinel-1/Sentinel-2 patch-pairs constituting the core of the \emph{SEN12MS} dataset, we add four-band MODIS land cover patches created from 2016 data at an upsampled pixel spacing of $10$\,m. The first of the provided bands contains land cover following the International Geosphere-Biosphere Programme (IGBP) classification scheme \cite{Loveland1997}, while the remaining bands contain the LCCS land cover layer, the LCCS land use layer, and the LCCS surface hydrology layer \cite{DiGregorio2005}. The schemes' classes are listed in Tab.~\ref{tab:lc_classes}. According to \cite{SullaMenashe2019}, the overall accuracies of the layers are about $67\%$ (IGBP), $74\%$ (LCCS land cover), $81\%$ (LCCS land use), and $87\%$ (LCCS surface hydrology), respectively. This should be kept in mind when using the land cover data as labels for training scene classification or semantic segmentation models, as these accuracies will constitute the upper bound of actually achievable predictive power -- even if validation accuracies of $100\%$ are reached. If the land cover information is not utilized as annotation, but as auxiliary data source, similar caution should be had.

\begin{table*}[!htb]
    \centering
    \begin{tabular}{ccccc}
    \toprule
        Class & IGBP value & LCCS LC value & LCCS LU value & LCCS SH value \\
        \cmidrule(r){1-1} \cmidrule(lr){2-2} \cmidrule(lr){3-3} \cmidrule(lr){4-4} \cmidrule(l){5-5}
        Evergreen Needleleaf Forests & 1 & 11 & -& -\\
        Evergreen Broadleaf Forests & 2 & 12 & - & -\\
        Deciduous Needleleaf Forests & 3 & 13 & - & -\\
        Deciduous Broadleaf Forests & 4 & 14 & - & -\\
        Mixed Broadleaf/Needleleaf Forests & - & 15 & - & -\\
        Mixed Broadleaf Evergreen/Deciduous Forests & - & 16 & - &-\\
        Mixed Forests & 5 & - & - &-\\
        Dense Forests & - & - & 10 & 10\\
        Open Forests & - & 21 & 20 & 20\\
        Sparse Forests & - & 22 & -&-\\
        Natural Herbaceous & - & - & 30&-\\
        Dense Herbaceous & - & 31 & -&-\\
        Sparse Herbaceous & - & 32 & -&-\\
        Shrublands & - & - & 40 & 40\\
        Closed (Dense) Shrublands & 6 & 41 & -&-\\
        Open (Sparse) Shrublands & 7 & 43 & -&-\\
        Shrubland/Grassland Mosaics & - & 42 & -&-\\
        Woody Savannas & 8 & - & -&-\\
        Savannas & 9 & - & -&-\\
        Grasslands & 10 & - & - & 30\\
        Permanent Wetlands & 11 & - & -&-\\
        Woody Wetlands & - & - & - & 27\\
        Herbaceous Wetlands & - & - & - & 50\\
        Herbaceous Croplands & - & - & 36&-\\
        Croplands & 12 & - & -&-\\
        Urban and Built-Up Lands & 13 & - & 9&-\\
        Cropland/Natural Vegetation Mosaics & 14 & - & -&-\\
        Forest/Cropland Mosaics & - & - & 25&-\\
        Natural Herbaceous/Croplands Mosaics & - & - & 35&-\\
        Tundra & - & - & - & 51\\
        Permanent Snow and Ice & 15 & 2 & 2 & 2\\
        Barren & 16 & 1 & 1 & 1\\
        Water Bodies & 17 & 3 & 3 & 3\\
        \bottomrule
    \end{tabular}
    \caption{MODIS land cover classes as represented by four different schemes: IGBP, LCCS land cover, LCCS land use, LCCS surface hydrology. The \emph{NoData} value is set to be 255 \protect\cite{sulla2018user}.}
    \label{tab:lc_classes}
\end{table*}

\section{Google Earth Engine for Data Preparation}
As for the \emph{SEN1-2} dataset, we have again utilized Google Earth Engine \cite{Gorelick2017} to generate a large-scale dataset of corresponding multi-sensor remote sensing image patches. While we basically used the same pipeline as described in \cite{Schmitt2018}, including the random sampling of ROIs for the meteorological seasons of the northern hemisphere, we added a more sophisticated mosaicking workflow for the generation of cloud-free short-term Sentinel-2 mosaics.

\subsection{Mosaicking of Cloud-Free Sentinel-2 Images}\label{sec:mosaic}
\begin{figure}[!tb]
\centering
\includegraphics[width=0.7\linewidth]{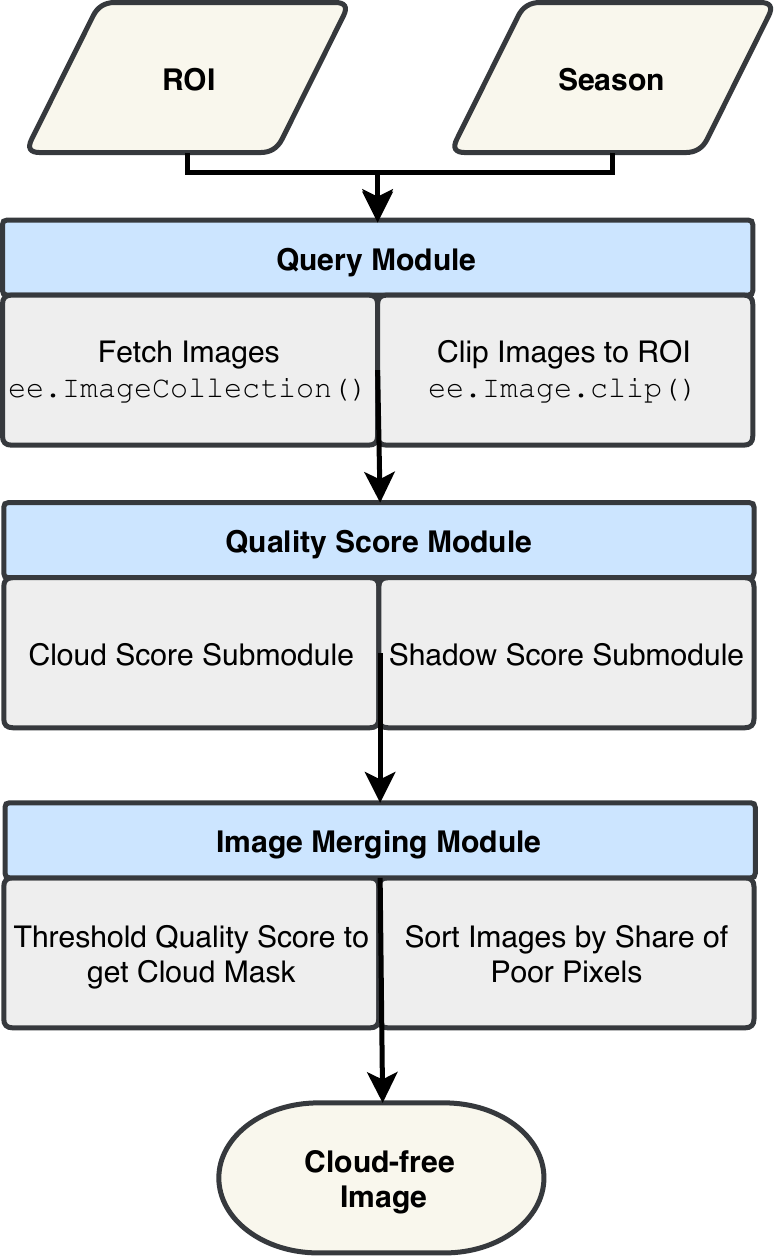}\caption{Workflow of the GEE-based procedure for cloud-free Sentinel-2 image generation presented in \protect\cite{Schmitt2019}.}\label{fig:CloudRemovalWorkflow}
\end{figure}

The general mosaicking workflow to produce cloud-free Sentinel-2 images for a given region of interest (ROI) and a specified time period is depicted in Fig.~\ref{fig:CloudRemovalWorkflow}. In essence, it consists of three main modules, which are carried out for every ROI. These ROIs result from two uniform random samplings over the landmasses of the Earth and the urban areas across the globe, respectively. 

While the procedure is described in detail in \cite{Schmitt2019}, a short summary of the three modules is as follows:
\begin{itemize}
    \item[(1)] The \emph{Query Module} for loading images from the catalogue. In this module, for the specified ROI all Sentinel-2 images available for a specified time period are selected.
    \item[(2)] The \emph{Quality Score Module} for the calculation of a quality score for each image. In this module, every pixel of each Sentinel-2 image is assigned a score that considers the likelihood it is affected either by clouds or by shadow. 
    \item[(3)] The \emph{Image Merging Module} for mosaicking of the selected images based on the meta-information generated in the preceding modules. First, the quality scores are thresholded to determine cloud and shadow masks for each image. Afterwards, the images are sorted by their amount of poor pixels. The best images are finally merged into a cloud-free mosaic.
\end{itemize} 
Since the Sentinel-1 images and the MODIS land cover data are not affected by clouds, in these cases no complicated mosaicking processes are required and the data are simply exported in a straight-forward manner later on.

\subsection{Data Export}
For \emph{SEN12MS} we have utilized the same random ROIs as for \emph{SEN1-2}. The same holds for the meteorological seasons as defined for the northern hemisphere.
After preparation of the Sentinel-1 images and the cloud-free Sentinel-2 mosaics for every ROI and season, we export them together with the land cover data at a scale of $10$\,m in the form of GeoTiffs. In this context, it has to be noted that the $10$\,m scale is defined at the equator by Google Earth Engine, which corresponds to an angular resolution of $0.0001^\circ$. 
This angular resolution leads to significantly smaller pixel widths for regions with latitudes deviating from $0^\circ$. We therefore used GDAL \cite{Warmerdam2008} to transform all exported data from the WGS84 lat/lon georeference to their local UTM coordinate representation, while resampling to actually square pixels of $10$\,m$\times10$\,m. 

\subsection{Data Curation}
\begin{figure}[!bth]
\centering
{\includegraphics[width=0.31\textwidth]{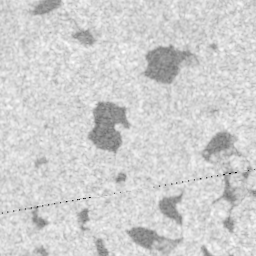}}
{\includegraphics[width=0.31\textwidth]{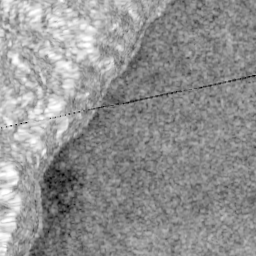}}
{\includegraphics[width=0.31\textwidth]{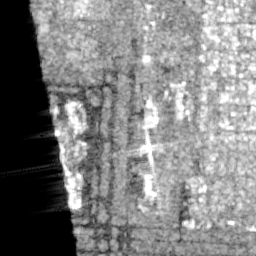}}
{\includegraphics[width=0.31\textwidth]{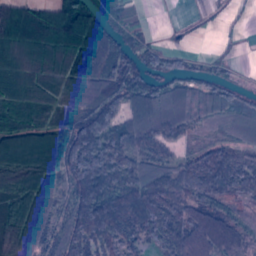}}
{\includegraphics[width=0.31\textwidth]{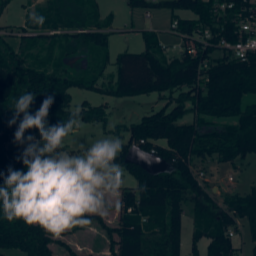}}
{\includegraphics[width=0.31\textwidth]{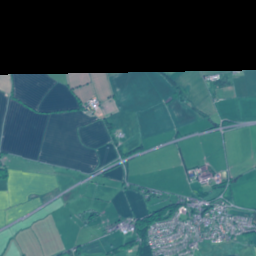}}
\caption{Examples for manually removed SAR (top row) and optical patches (bottom) row).}
\label{fig:removedPatches}
\end{figure}

\begin{figure*}[]
    \centering
    \includegraphics[width=0.9\linewidth]{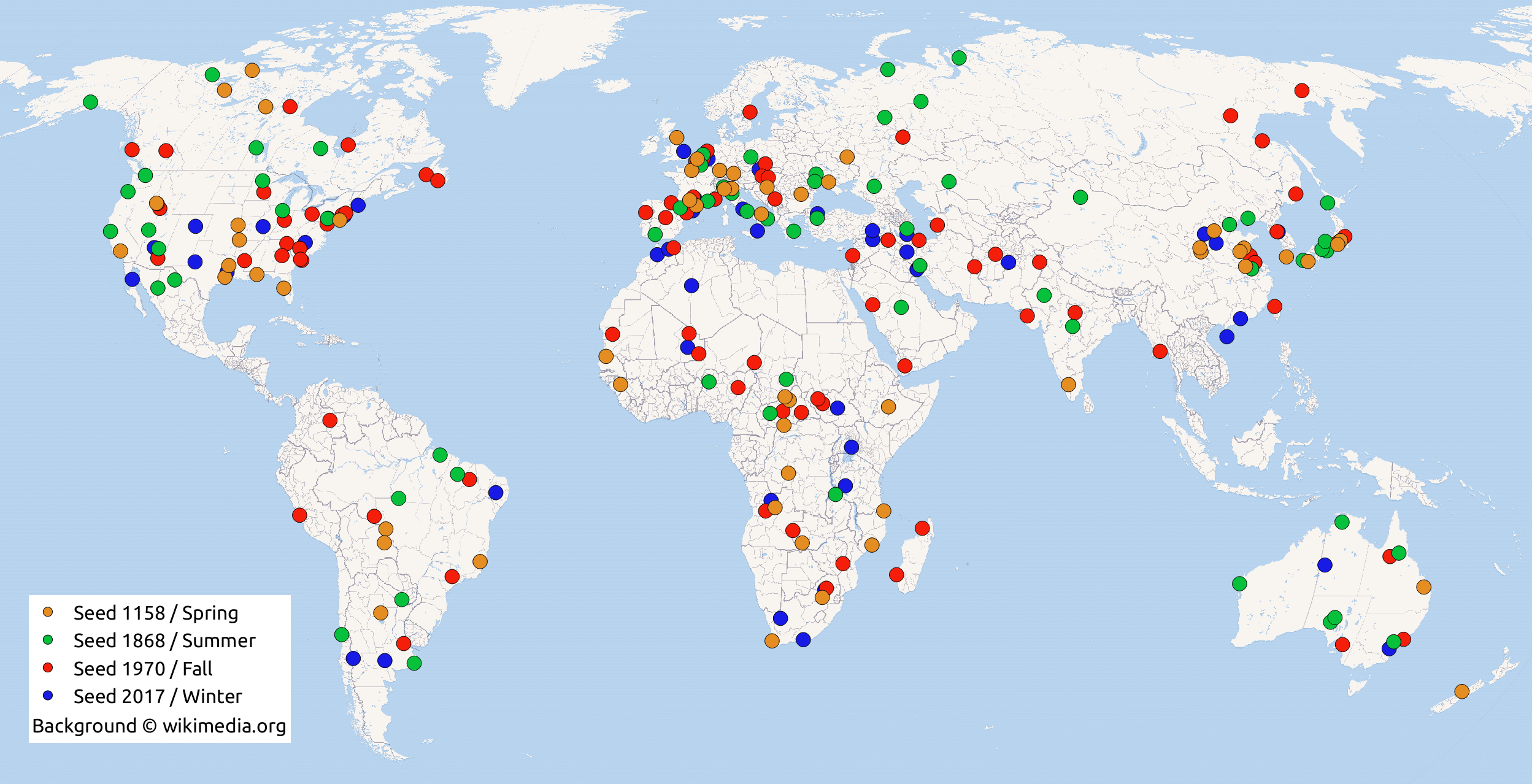}
    \caption{Final distribution of the ROIs over all inhabited land masses of the Earth.}
    \label{fig:WorldMap}
\end{figure*}

After the export of the data from the GEE servers to local storage, we followed an inspection protocol similar to the one proposed in our previous work \cite{Schmitt2018}: First, each triplet of full scene images was converted to a visually perceivable format (gray-scale images for Sentinel-1 and MODIS Land Cover, RGB images for Sentinel-2) and displayed to a remote sensing expert. If either of the three images contained very large no-data areas, large non-detected clouds, or strong artifacts resulting from the cloud-adaptive mosaicking, the triplet was discarded. After this first inspection, only $252$ out of the originally downloaded $600$ scenes were kept in the dataset. 
These remaining scenes were then tiled into patches of $256\times 256$ pixels in size. Again, we have implemented a stride of $128$ pixels, resulting in an overlap between adjacent patches of $50\%$. 
We think, a $50\%$ overlap is the ideal trade-off between patch independence and maximization of the number of samples. After the tiling, $216{,}596$ patch triplets were available for a second inspection.
In this second inspection, all patches were again visually inspected by remote sensing experts in order to avoid patches containing artefacts or distortions, e.g. no data areas, clouds, or jet streams. Some examples for patch types that were discarded in this step are displayed in Fig.~\ref{fig:removedPatches}. After this final inspection step, a total of $180{,}662$ patch triplets remained, which comprise the final \emph{SEN12MS} dataset. The locations of the final ROI scene locations are displayed in Fig.~\ref{fig:WorldMap}.

\section{The SEN12MS Dataset}\label{sec:dataset}
\begin{figure}[!tbh]
\centering
{\includegraphics[width=0.31\textwidth]{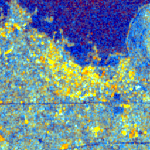}}
{\includegraphics[width=0.31\textwidth]{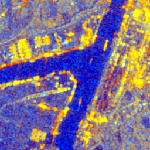}}
{\includegraphics[width=0.31\textwidth]{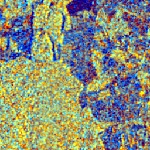}}
{\includegraphics[width=0.31\textwidth]{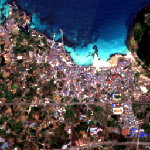}}
{\includegraphics[width=0.31\textwidth]{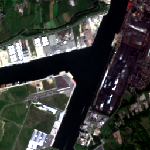}}
{\includegraphics[width=0.31\textwidth]{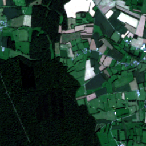}}
{\includegraphics[width=0.31\textwidth]{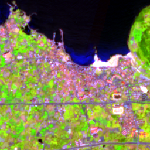}}
{\includegraphics[width=0.31\textwidth]{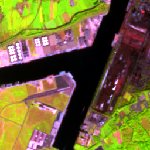}}
{\includegraphics[width=0.31\textwidth]{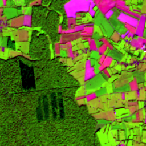}}
{\includegraphics[width=0.31\textwidth]{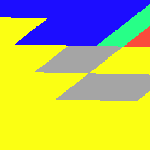}}
{\includegraphics[width=0.31\textwidth]{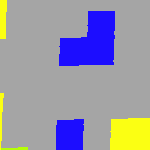}}
{\includegraphics[width=0.31\textwidth]{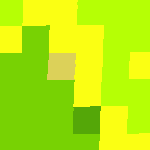}}
{\includegraphics[width=0.31\textwidth]{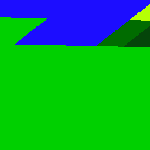}}
{\includegraphics[width=0.31\textwidth]{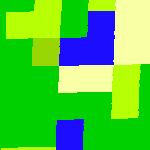}}
{\includegraphics[width=0.31\textwidth]{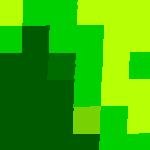}}
\caption{Images extracted from $3$ example patch triplets. Each column shows (from top to bottom): False color Sentinel-1 SAR (R: VV, G: VH, B: VV/VH), Sentinel-2 RGB, Sentinel-2 SWIR, IGBP Land cover, LCCS Land cover. Note that while the GSD of all patches is upsampled to $10$\,m, the actual resolution varies from $10$\,m (Sentinel-2 RGB) to $500$\,m (land cover).}
\label{fig:goodTriplets}
\end{figure}

The final \emph{SEN12MS} dataset contains $180{,}662$ patch triplets (Sentinel-1 dual-pol SAR, Sentinel-2 multi-spectral, MODIS land cover) in the form of multi-channel GeoTiff images and requires $421.3$\,GiB of storage. Some patch triplet examples are shown in Fig. \ref{fig:goodTriplets} to give an impression of the rich and versatile information contained in the dataset.

\subsection{Structure of the Final Dataset}
\begin{figure*}[h]
    \centering
    \includegraphics[width=0.6\textwidth]{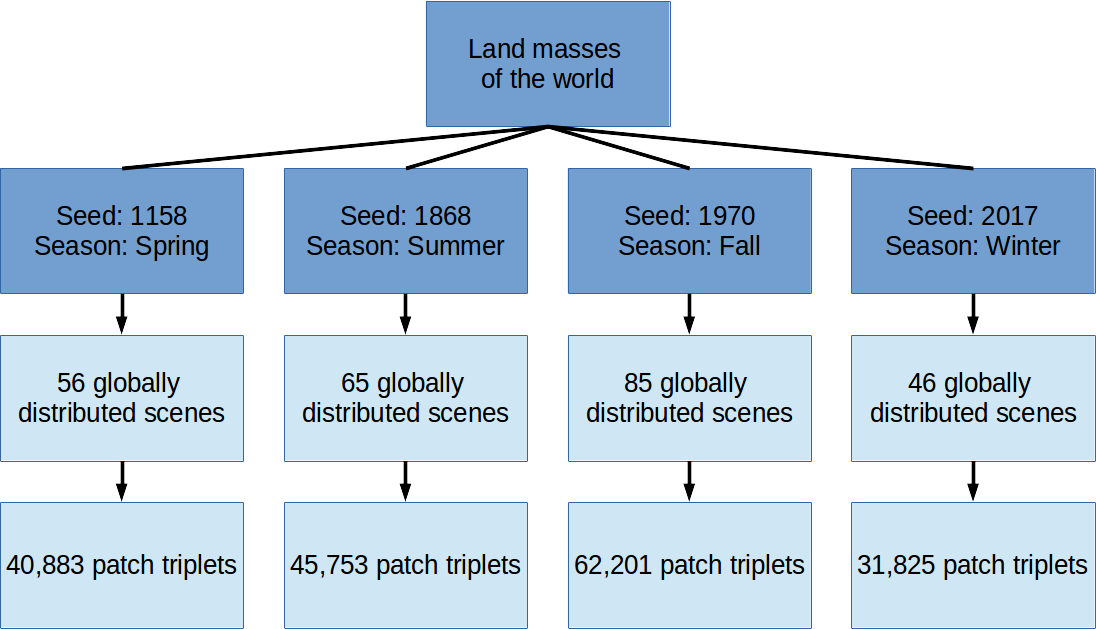}
    \caption{Tree structure of the final dataset.}
    \label{fig:DatasetStructure}
\end{figure*}

As mentioned in Section~\ref{sec:mosaic}, the dataset is based on randomly sampled regions of interest, resulting from four different seed values: \longword{1158}, \longword{1868}, \longword{1970}, and \longword{2017}. These four different seed values are related to the four meteorological seasons defined for the northern hemisphere: winter (1 December 2016 to 28 February 2017), spring (1 March 2017 to 30 May 2017), summer (1 June 2017 to 31 August 2017), and fall (1 September 2017 to 30 November 2017). This leads to a tree-like structure of the dataset into four branches: \longword{ROIs1158\_spring}, \longword{ROIs1868\_summer},
\longword{ROIs1970\_fall}, and \longword{ROIs2017\_winter}. Each of those branches again is divided into several sub-branches corresponding to the individual ROIs (or scenes, respectively) derived from the corresponding random number seed. The full dataset structure is depicted in Fig.~\ref{fig:DatasetStructure}.

The Sentinel-1 data can be recognized by the abbreviation \longword{s1}, the Sentinel-2 data by \longword{s2}, and the MODIS land cover data by \longword{lc}; the individual patches can be identified by the token \longword{pXXX} where \longword{XXX} denotes a unique identifier number per patch. Thus, the file naming convention follows the following scheme:
\longword{ROIsSSSS\_SEASON\_DD\_pXXX.tif}, where \longword{SSSS} denotes the seed value, \longword{SEASON} denotes the meteorological season as defined for the northern hemisphere, \longword{DD} denotes the data identifier, and \longword{XXX} denotes the patch identifier.

Of course, we are aware that the seasonal structuring of the dataset is only of little semantic worth since we have taken the seasons of the northern hemisphere as a reference. To allow end-users a sub-structuring of the dataset taking semantically meaningful seasons into account, we provide the file \longword{seasons.csv} with the metadata of the dataset. It declares, which scenes actually were acquired in spring, summer, winter, and fall from a climatic point of view.

While we forgo to define a fixed train/test split, we think this can easily be achieved by end-users considering their individual needs: Deterministic splits into disjunct training and test sets can be achieved via the meteorological seasons, or via the individual ROIs.

\subsection{Dataset Availability}
The \emph{SEN12MS} dataset is shared under the open access license CC-BY and available for download at a persistent link provided by the library of the Technical University of Munich (TUM): \url{https://mediatum.ub.tum.de/1474000}.
This paper must be cited when the dataset is used for research purposes. 

\section{Application to Land Cover Mapping}\label{sec:experiments}
\begin{figure*}[!bth]
\centering
\subfigure[][]{
\includegraphics[width=0.22\textwidth]{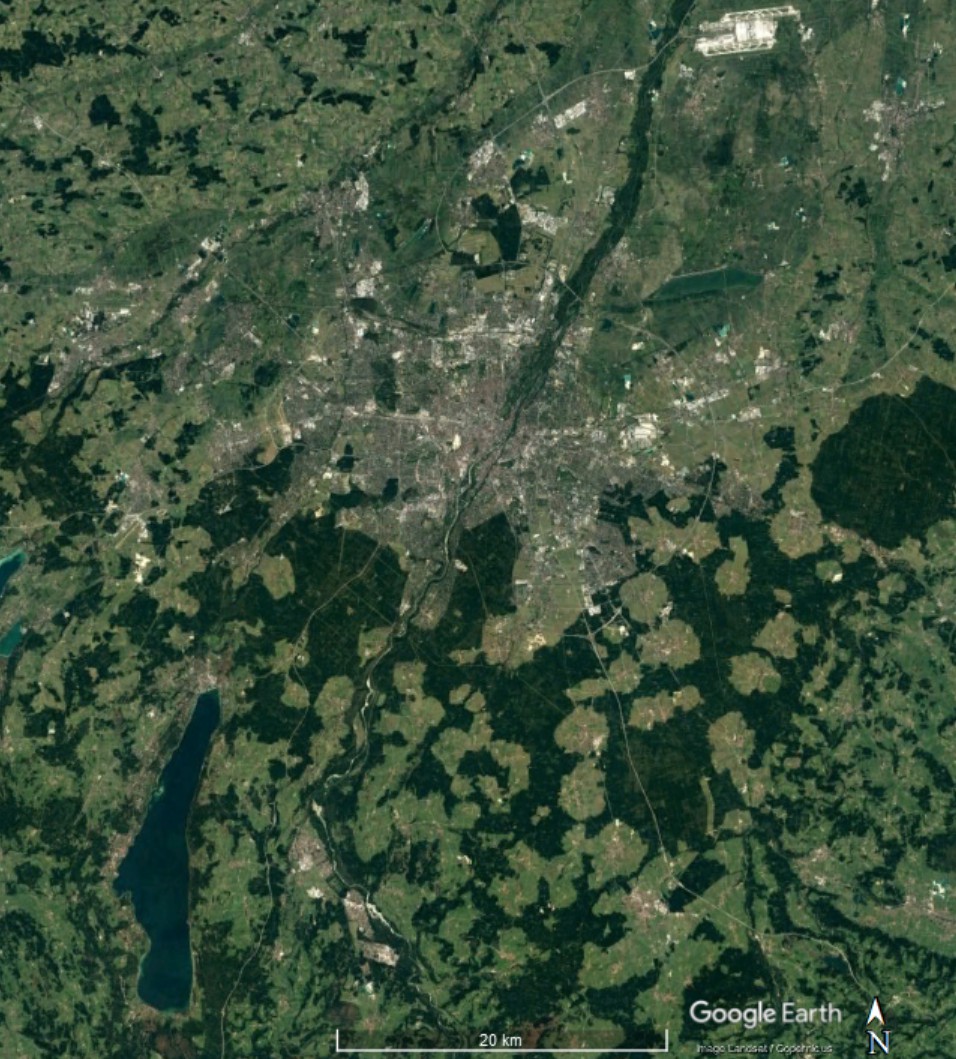}
}
\subfigure[][]{
\includegraphics[width=0.22\textwidth]{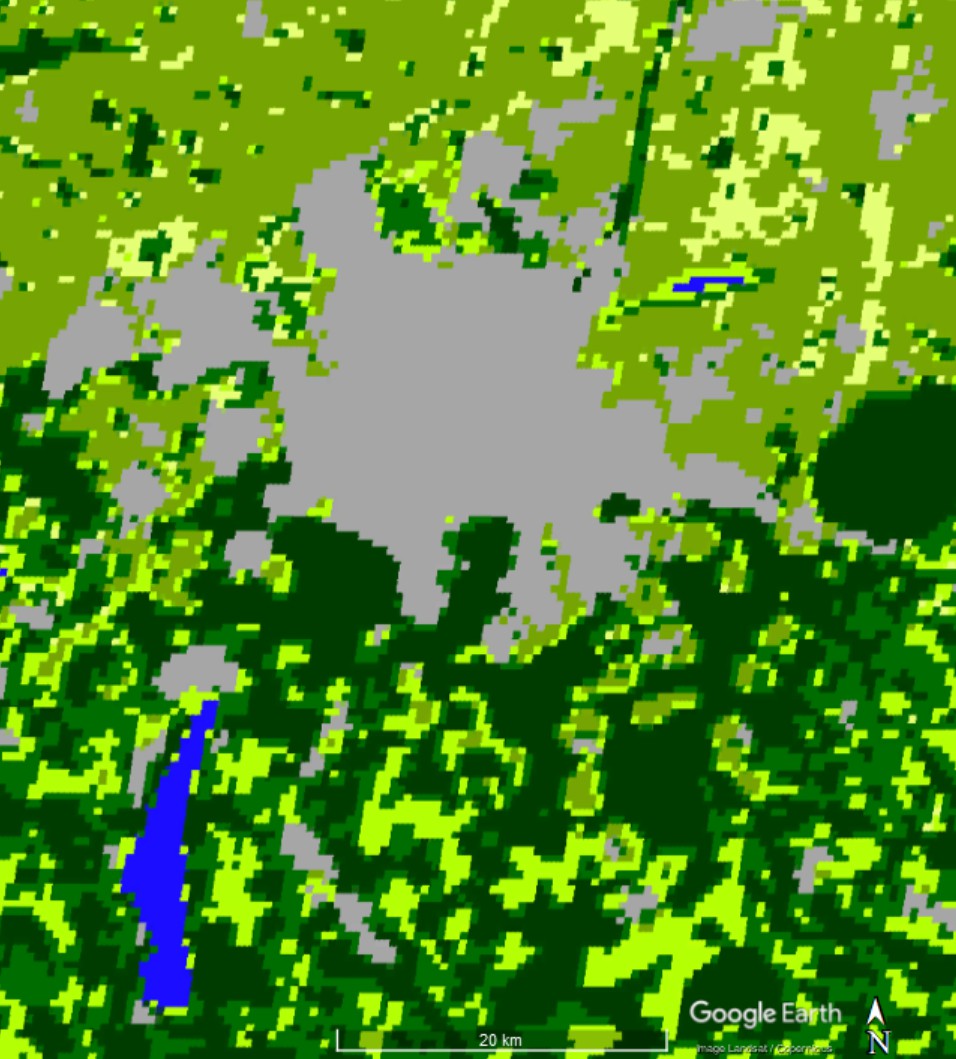}
}
\subfigure[][]{
\includegraphics[width=0.22\textwidth]{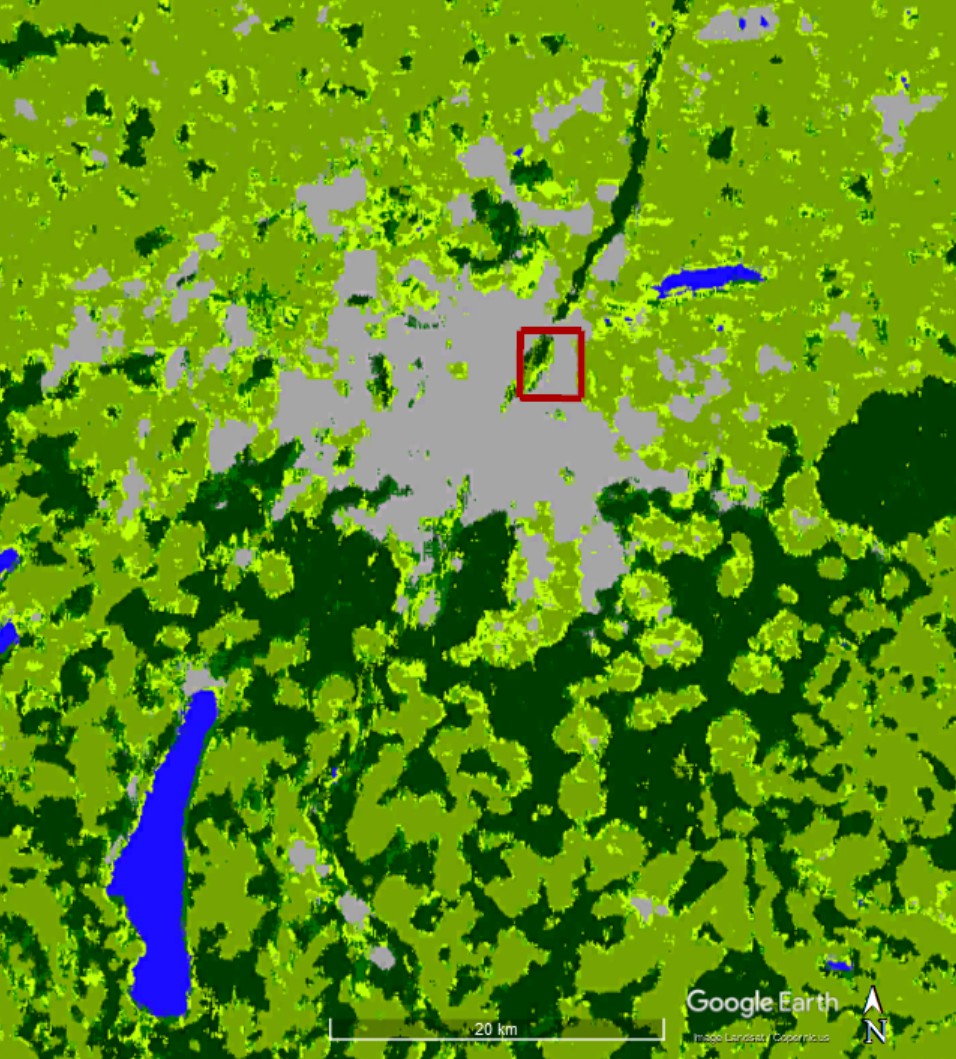}
}
\subfigure[][]{
\includegraphics[width=0.22\textwidth]{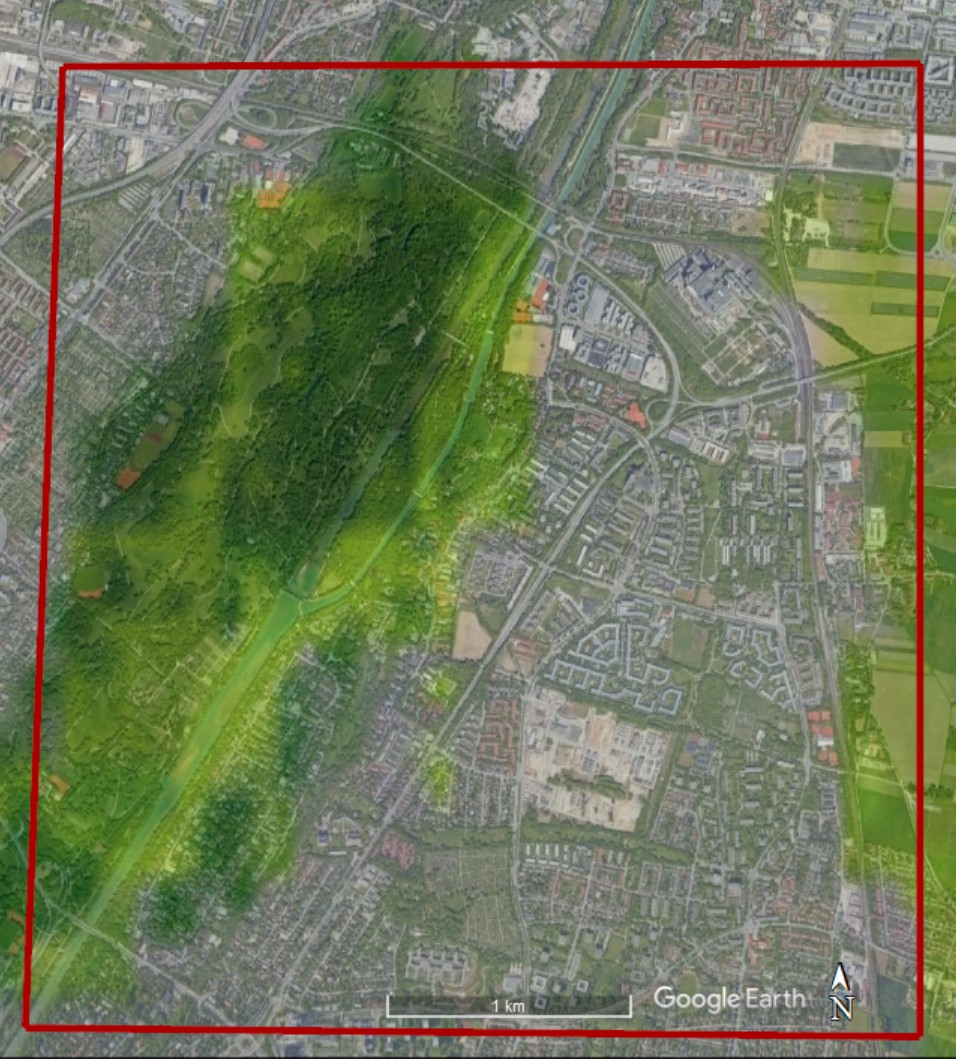}
}

\subfigure[][]{
\includegraphics[width=0.22\textwidth]{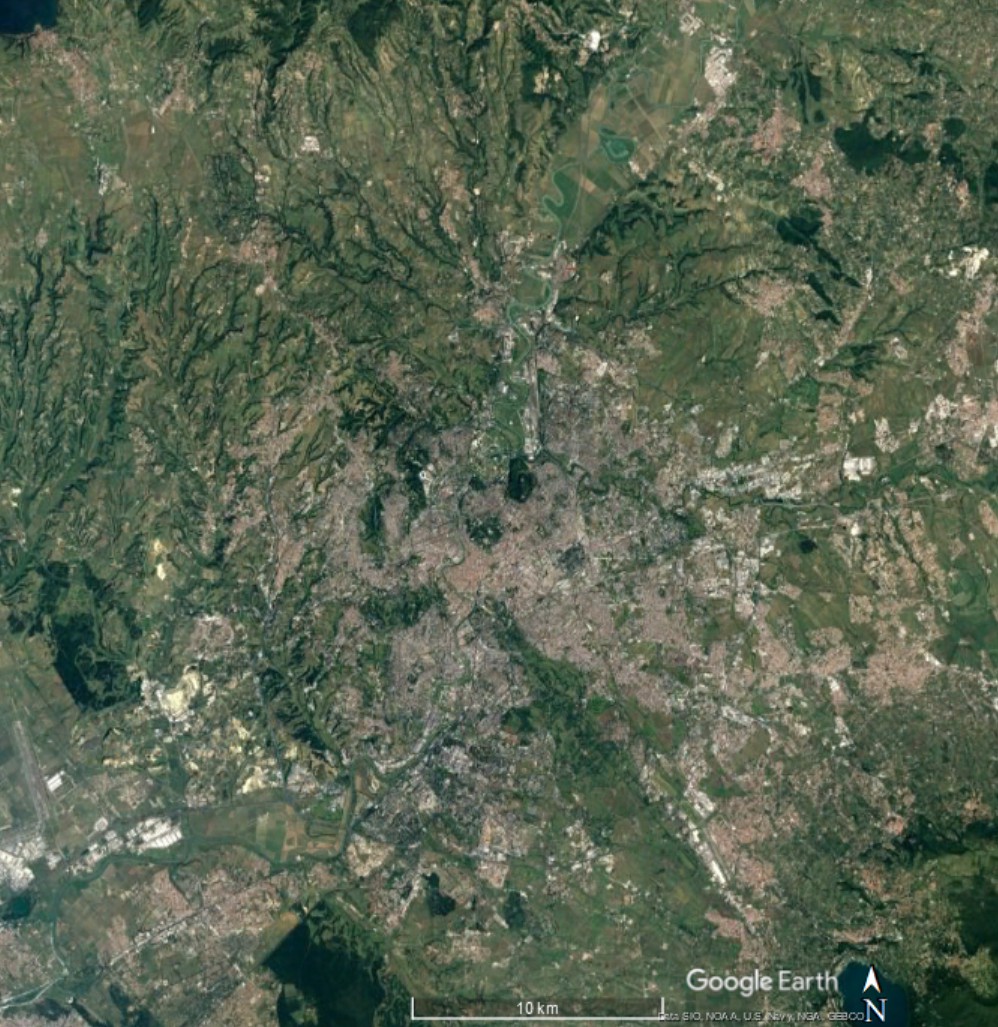}
}
\subfigure[][]{
\includegraphics[width=0.22\textwidth]{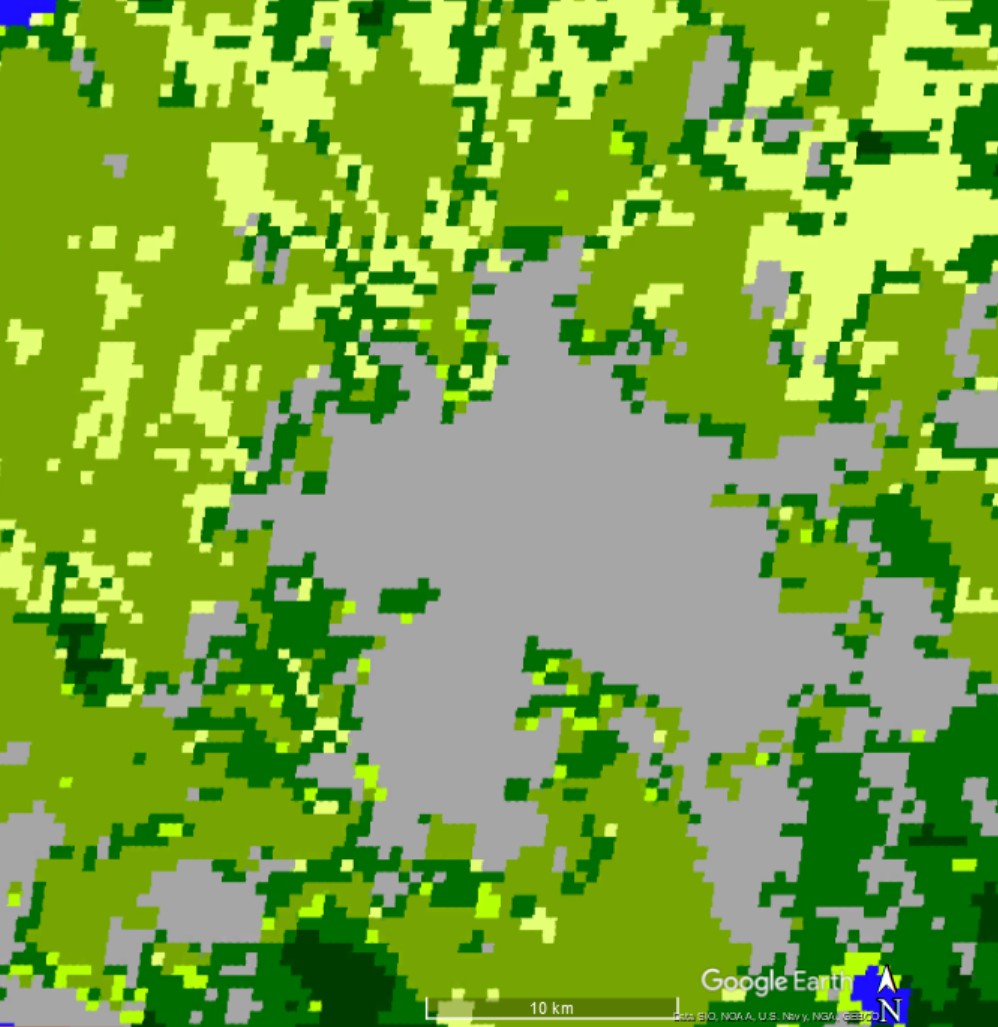}
}
\subfigure[][]{
\includegraphics[width=0.22\textwidth]{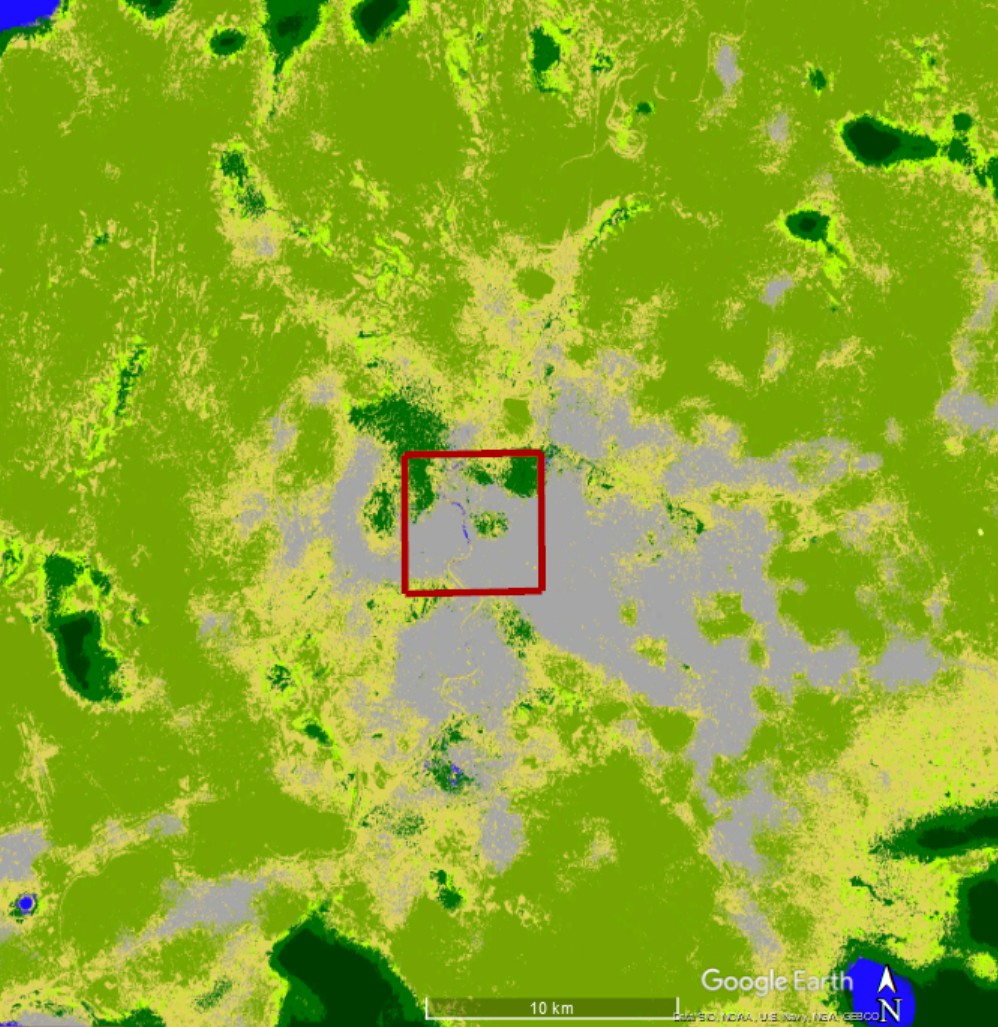}
}
\subfigure[][]{
\includegraphics[width=0.22\textwidth]{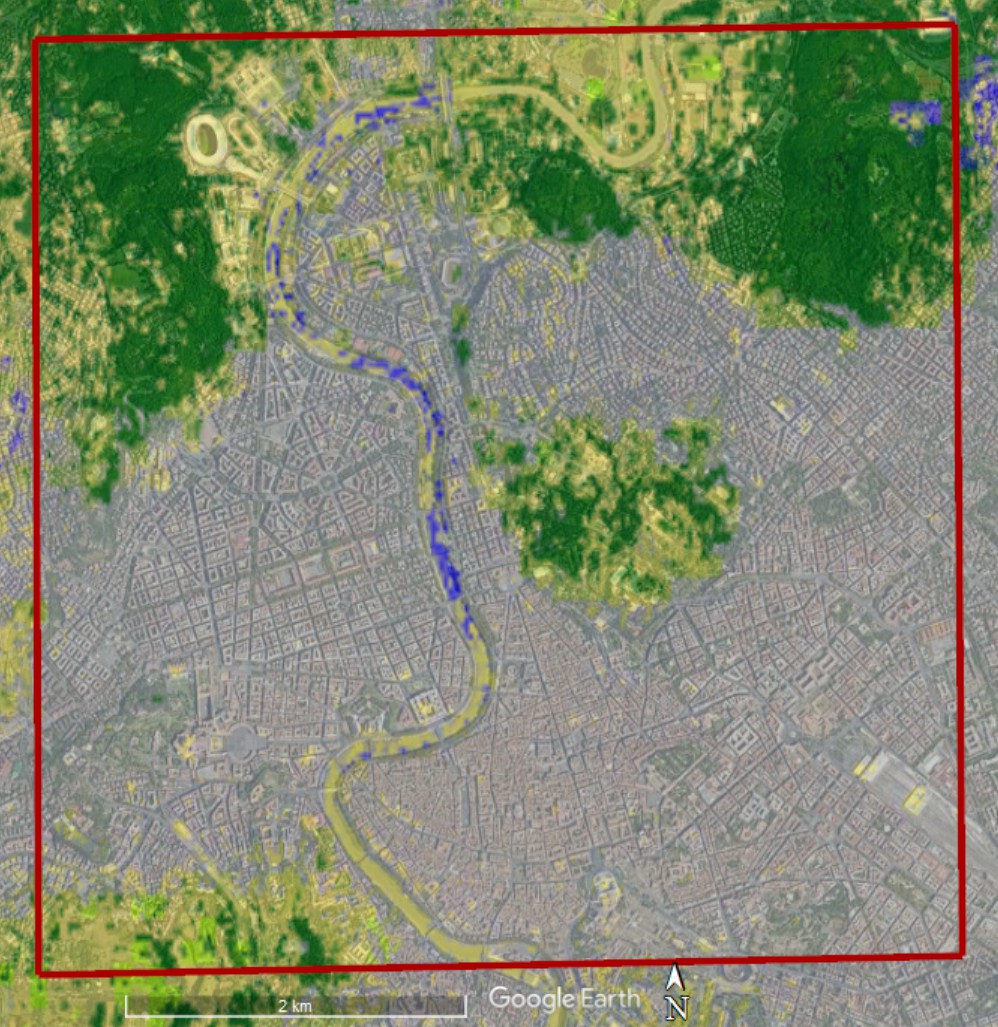}
}
\begin{tikzpicture}
\pgfplotsset{
legend style={cells={anchor=west}, draw=none,column sep=1ex, nodes={scale=0.7, transform shape}}
} 
\begin{customlegend}[legend columns=4]
\addlegendimage{lu1, only marks, mark=square*}
\addlegendentry{Barren (LCCS 1)}
\addlegendimage{lu2, only marks, mark=square*}
\addlegendentry{Permanent Snow and Ice (LCCS 2)}
\addlegendimage{lu3, only marks, mark=square*}
\addlegendentry{Water Bodies (LCCS 3)}
\addlegendimage{lu4, only marks, mark=square*}
\addlegendentry{Urban and Built-Up Lands (LCCS 9)}
\addlegendimage{lu5, only marks, mark=square*}
\addlegendentry{Dense Forests (LCCS 10)}
\addlegendimage{lu6, only marks, mark=square*}
\addlegendentry{Open Forests (LCCS 20)}
\addlegendimage{lu7, only marks, mark=square*}
\addlegendentry{Forest/Cropland Mosaics (LCCS 25)}
\addlegendimage{lu8, only marks, mark=square*}
\addlegendentry{Natural Herbaceous (LCCS 30)}
\addlegendimage{lu9, only marks, mark=square*}
\addlegendentry{Natural Herbaceous/Croplands Mosaics (LCCS 35)}
\addlegendimage{lu10, only marks, mark=square*}
\addlegendentry{Herbaceous Croplands (LCCS 36)}
\addlegendimage{lu11, only marks, mark=square*}
\addlegendentry{Shrublands (LCCS 40)}
\end{customlegend}
\end{tikzpicture}
\caption{Examples from Munich (top) and Rome (bottom) for the predictive power of the dataset: (a) and (e) Optical images extracted from Google Earth, (b) and (f) MODIS-derived LCCS land use map with 500m resolution, (c) predicted LCCS land use map with 100m resolution using classification network, (g) predicted LCCS land use map with 10m resolution using semantic segmentation network, (d) and (h) zoom-in of the red rectangles in the respective predicted LCCS land use maps. While the area in the red rectangle is completely defined as urban in the original MODIS-derived product, more details are visible in the predicted results.}
  \label{fig:map}
\end{figure*}

In order to provide an example for the usefulness of the dataset with regard to the development of land cover classification solutions, we have trained two state-of-the-art deep convolutional neural network architectures for predicting LCCS land use classes (cf. Tab.~\ref{tab:baselineNets}). 
\begin{table}[tbh]
    \centering
    \begin{tabular}{ p{0.28\linewidth} p{0.28\linewidth} p{0.28\linewidth}}
    \toprule
     ~ & ResNet-110 & DenseNet\\
      \cmidrule(r){2-2} \cmidrule(l){3-3}
        \textit{Upsampling type} & n/a & Deconvolution\\
        \textit{Input size} & $64\times64\times10$ & $256\times256\times10$\\
        \textit{Batch size} & $16$ & $4$\\
        \textit{Loss} & \multicolumn{2}{c}{Categorical cross-entropy} \\
        \textit{Optimizer} & \multicolumn{2}{c}{Nesterov Adam}\\
        \textit{Initial LR} & 0.0005 & 0.0001\\
        \textit{LR schedule} & \multicolumn{2}{c}{ReduceOnPlateau}\\
        \bottomrule
    \end{tabular}
    \caption{Baseline network training configurations.}
    \label{tab:baselineNets}
\end{table}
The first network, ResNet-110 \cite{he2016deep}, was designed for image classification, i.e. for assigning a single class label to the input image. For the presented baseline experiment, we cut images of $64\times64$ pixels from the Sentinel-2 samples from the \emph{summer} subset, and used the following ten bands as channel information: B2 (Blue), B3 (Green), B4 (Red), B8 (Near-infrared), B5 (Red Edge 1), B6 (Red Edge 2), B7 (Red Edge 3), B8a (Red Edge 4), B11 (Short-wavelength infrared 1), and B12 (Short-wavelength infrared 2). We then used the majority LCCS land use class from each of the $64\times64$ patches as scene label for that patch. Due to the unconventional multi-channel configuration of the data, we trained the network from scratch rather than relying on any pre-trained weights. In order to test the predictive power of this network, we applied it to the area of the city of Munich, with the test image being pre-processed with the same GEE-based procedure as described in Section~\ref{sec:mosaic}. The resulting land use map is shown in Fig.~\ref{fig:map} (top row), and was created by sliding the ResNet-110 on an input Sentinel-2 image with a stride of $10$ (leading to an output pixel spacing of $100$\,m). The overall accuracy (OA), average accuracy (AA) and Kappa coefficient calculated based on a grid of manually annotated control points can be seen in Tab.~\ref{tab:resultsMunich}. It has to be noted that for sake of this evaluation, we combined the classes \emph{Open Forests} (LCCS 20) and \emph{Forest/Cropland Mosaic} (LCCS 25) to a joint \emph{Open Forests} class, and \emph{Natural Herbaceous} (LCCS 30) with \emph{Herbacoeus Croplands} (LCCS 36) and \emph{Natural Herbaceous/Croplands Mosaic} (LCCS 35) to a simple \emph{Herbaceous} class, as it is very difficult for human annotators to distinguish those classes by their only subtle differences.

The second network we used was the fully convolutional DenseNet for semantic segmentation \cite{jegou2017one}, aiming at assigning a class label to every pixel of the input image. Here, we used full-sized Sentinel-2 10-band patches (i.e. $256\times256$\,pixels) as input and processed the city of Rome for test purposes. The result is also depicted in Fig.~\ref{fig:map}, and the accuracy metrics, calculated analogue to the Munich case, are described in Tab.~\ref{tab:resultsRome}.

\begin{table}[htbp]
  \centering
    \begin{tabular}{lccc}
    \toprule
    & OA    & Kappa & AA   \\
    \cmidrule(r){2-2}\cmidrule(lr){3-3}\cmidrule(l){4-4}
    \textit{MODIS} & 53.0\% & 0.38  & 37.5\% \\
    \textit{Predicted} & 65.1\% & 0.51  & 43.6\% \\
    \bottomrule
    \end{tabular}%
    \caption{Accuracy metrics of the LCCS maps for the Munich scene. The predicted result was achieved by patch classification based on a ResNet-110 CNN. The evaluation is based on manually labeled reference points.}
  \label{tab:resultsMunich}%
\end{table}%
\begin{table}[htbp]
  \centering
    \begin{tabular}{lccc}
    \toprule
    & OA    & Kappa & AA   \\
    \cmidrule(r){2-2}\cmidrule(lr){3-3}\cmidrule(l){4-4}
    \textit{MODIS} & 56.0\% & 0.37  & 33.7\% \\
    \textit{Predicted} & 63.3\% & 0.44  & 39.4\%  \\
    \bottomrule
    \end{tabular}%
    \caption{Accuracy metrics of the LCCS maps for the Rome scene. The predicted result was achieved by semantic segmentation based on a DenseNet CNN. The evaluation is based on manually labeled reference points.}
  \label{tab:resultsRome}%
\end{table}%

By comparing the resulting maps to the original MODIS-derived LCCS land use map, as well as an image extracted from Google Earth, it can be seen that in both cases the resolution of the map was successfully enhanced so that more details can be retrieved, while an overall agreement between the low-resolution MODIS map and the corresponding predicted high-resolution result is preserved. It has to be highlighed that both test areas are not contained in the \emph{SEN12MS} dataset, so that the results can serve as a first indicator of the strong generalization capability provided by the versatility of the dataset. This holds even more so since only a small subset of the dataset (namely the patches of the \emph{summer} season) has been used for training the networks used in the experiments. This impression is also confirmed by the independently evaluated accuracy metrics. We expect that by exploiting the whole dataset, i.e. both sensor modalities, all ROIs and all seasons, powerful classifiers for large-scale mapping applications can be developed. Besides that, it has to be mentioned that training on low-resolution labels of course creates a form of label noise. Using specifically adapted strategies aiming at so-called label super-resolution \cite{Malkin2019} might be able to provide even better results.

\section{Discussion}
As can be seen from Tab.~\ref{tab:datasets}, \emph{SEN12MS} is among the five largest datasets when only the sheer number of image patches is considered. However, in the end, \emph{SEN12MS} contains a lot more data and is consequently much bigger than its competitors: First and foremost, its patches are of size $256 \times 256$ pixels instead of just $28 \times 28$ (SAT-4/6) or $120 \times 120$ (BigEarthNet) pixels. Besides, the spectral information content is also much higher, as \emph{SEN12MS} contains full multi-spectral Sentinel-2 imagery and, in addition, dual-polarimetric Sentinel-1 SAR data, while most other datasets -- besides EuroSAT, SEN1-2, and BigEarthNet -- only contain color imagery with 3 or 4 bands. Last, but not least, it has to be mentioned that \emph{SEN12MS} is the most versatile dataset regarding scene distribution, as it covers all regions of the Earth over all meteorological seasons, while most of the other datasets are restricted to fairly small areas (e.g. Brazilian Coffee Scenes or USGS SIRI-WHU), individual countries (e.g. SAT-4/5 or DeepGlobe - Road Extraction), or a single continent (e.g. BigEarthNet). 

Wile this versatility is the major strength of \emph{SEN12MS}, it also poses the greatest challenge: In comparison to, e.g., ImageNet, which also provides a lot of versatility and thus tries to model the whole world, the number of samples in \emph{SEN12MS} is still fairly small. 
As can be seen from the example results provided in Section~\ref{sec:experiments}, the dataset nevertheless seems to hold the potential to train powerful, well-generalizing models even under data-scarce training situations. Besides, we believe that further benefit will arise from combining the existing datasets, e.g. BigEarthNet, with \emph{SEN12MS} in the frame of transfer learning to enlarge the amount of data a model can learn patterns from.  

\section{Summary and Conclusion}
With this paper, we publish the \emph{SEN12MS} dataset, which contains $180{,}662$ triplets of Sentinel-1 dual-polarimetric SAR data, Sentinel-2 multi-spectral images, and MODIS-derived land cover maps. With its large patch size, its global scene distribution, and its wealth of versatile remote sensing information, it can be considered to be the largest remote sensing dataset available to date. We hope it will foster the development of well-generalizing machine learning models for a more sophisticated automatic analysis of Sentinel satellite data.

\section*{ACKNOWLEDGEMENTS}\label{Acknowledgments}
This work is jointly supported by the Helmholtz Association under the framework of the Young Investigators Group SiPEO (VH-NG-1018), the German Research Foundation (DFG, grant SCHM 3322/1-1), and the European Research Council (ERC) under the European Union’s Horizon 2020 research and innovation programme (grant agreement ERC-2016-StG-714087, Acronym: So2Sat).

{
	\begin{spacing}{0.8}
		\bibliography{references} 
	\end{spacing}
}

\end{document}